\documentclass[10pt,twocolumn,letterpaper]{article}

\usepackage[pagenumbers]{cvpr} 

\usepackage{multirow}
\usepackage{colortbl}
\usepackage{pifont}
\usepackage{marvosym}
\newcommand{\pub}[1]{{\color{gray}{\tiny{[{#1}]}}}}

\definecolor{cvprblue}{rgb}{0.21,0.49,0.74}
\usepackage{lineno}
\usepackage[pagebackref,breaklinks,colorlinks,citecolor=cvprblue]{hyperref}

\def\paperID{9246} 
\def\confName{CVPR}
\def\confYear{2025}

\title{Region-Aware Text-to-Image Generation via Hard Binding and Soft Refinement}

\author{Zhennan Chen$^1$$^{\star}$ ~ Yajie Li$^1$$^{\star}$ ~ Haofan Wang$^{2,3}$ ~ Zhibo Chen$^3$ ~ Zhengkai Jiang$^4$ \\
	Jun Li$^1$ ~ Qian Wang$^5$ ~ Jian Yang$^1$ ~ Ying Tai$^1$\textsuperscript{\Letter} \\
	$^1$Nanjing University  ~~  $^2$InstantX Team~~ $^3$Liblib AI ~~ $^4$HKUST ~~ $^5$China Mobile  \\
	{\small \textcolor{magenta}{\url{https://github.com/NJU-PCALab/RAG-Diffusion}}} \vspace{-4mm}
}

\begin{document}
\twocolumn[{
\renewcommand\twocolumn[1][]{#1}
\maketitle
\begin{center}
    \captionsetup{type=figure}
    \includegraphics[width=0.98\textwidth]{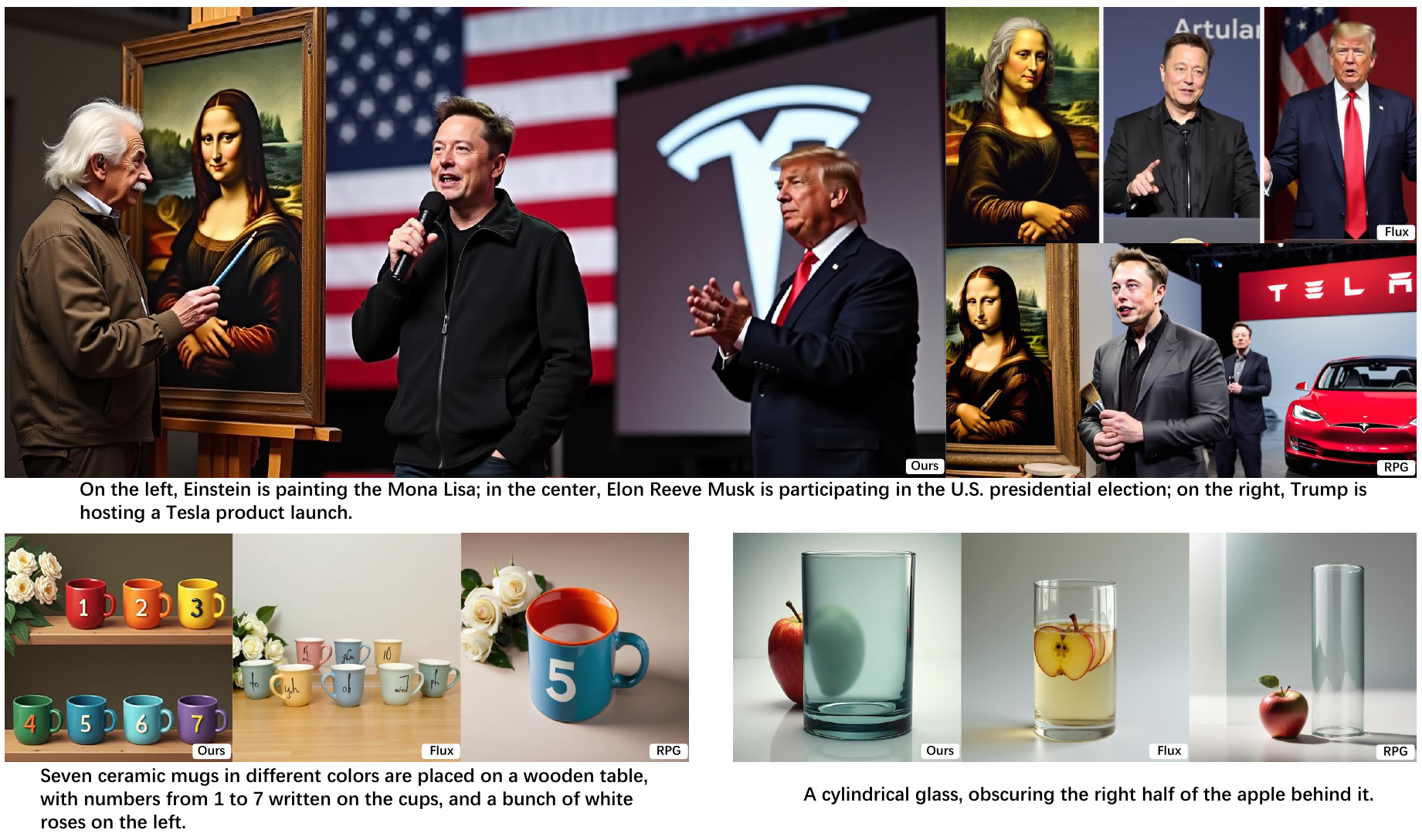}
    \vspace{-3.4mm}
    \caption{RAG decouples a raw prompt into regional prompts, processes different regions separately and strengthens the interaction between adjacent regions. It enables \textit{precise control over object relationship, action and attributes}, achieving more harmonious and consistent complex compositional generation compared to competing models like Flux and RPG.}
    \label{fig:teaser}
\end{center}
}]

\begin{abstract}
\vspace{-8mm}

Regional prompting, or compositional generation, which enables fine-grained spatial control, has gained increasing attention for its practicality in real-world applications. However, previous methods either introduce additional trainable modules, thus only applicable to specific models, or manipulate on score maps within cross-attention layers using attention masks, resulting in limited control strength when the number of regions increases. To handle these limitations, we present \textbf{RAG}, a \textbf{R}egional-\textbf{A}ware text-to-image \textbf{G}eneration method conditioned on regional descriptions for precise layout composition. RAG decouple the multi-region generation into two sub-tasks, the construction of individual region (\textbf{Regional Hard Binding}) that ensures the regional prompt is properly executed, and the overall detail refinement (\textbf{Regional Soft Refinement}) over regions that dismiss the visual boundaries and enhance adjacent interactions. Furthermore, RAG novelly makes repainting feasible, where users can modify specific unsatisfied regions in the last generation while keeping all other regions unchanged, without relying on additional inpainting models. Our approach is tuning-free and applicable to other frameworks as an enhancement to the prompt following property. Quantitative and qualitative experiments demonstrate that RAG achieves superior performance over attribute binding and object relationship than previous tuning-free methods. 

\end{abstract}\vspace{-12mm}    
\section{Introduction}
\label{sec:intro}

Recent advancements in diffusion models \cite{sohl2015deep, ho2020denoising, song2020score, peebles2023scalable,ramesh2022hierarchical,saharia2022photorealistic,yu2022scaling,xie2024sana,ye2023ip,wang2024instantid,wang2024instantstyle,wang2024instantstyleplus} have substantially enhanced the aesthetic appeal and prompt adherence in text-to-image synthesis. The prevailing trend sees the denoising architecture shifting from UNet \cite{ronneberger2015u} to the Diffusion Transformer (DiT) \cite{peebles2023scalable}, which excels in scaling with large datasets. Transformer-based diffusion models like PixArt-$\alpha$ \cite{chen2023pixart}, Stable Diffusion 3/3.5 \cite{esser2024scaling}, and Flux.1 \cite{flux} have set a new benchmark, surpassing the quality of earlier UNet-based models such as Stable Diffusion 1.5 \cite{rombach2022high} and SDXL \cite{podell2023sdxl}. Furthermore, employing more robust text encoders, including T5-XXL \cite{raffel2020exploring}, has demonstrated the ability to render visual text and significantly enhance prompt adherence. Some innovative studies even leverage large language models (LLMs) for text representation, with examples like Kolors \cite{kolors} utilizing GLM \cite{du2021glm} and Playground V3 \cite{liu2024playground} employing Llama3-8B \cite{dubey2024llama}. Despite this significant progress in generating high-quality images from prompts, achieving precise fine-grained spatial control remains elusive. In essence, current generative models still struggle with comprehending the quantity and spatial arrangement of objects.

To address these limitations, the concept of regional prompting, also known as regional control, regional grounding, or composition generation, has emerged. Unlike providing a single global description, achieving fine-grained region-controllable generation requires users to supply not only the spatial location (\eg, a segmentation mask or bounding box) but also a corresponding description for each region. Several approaches have been proposed under this setting, broadly falling into two categories: tuning-based and tuning-free. For tuning-based methods, they often necessitate the training of an additional module to handle explicit conditions like bounding boxes. For example, GLIGEN \cite{li2023gligen} integrates regional inputs into new trainable layers through a gated mechanism, where each grounding token is a combination of the semantics of the grounded entity and its spatial location. Similarly, InstanceDiffusion \cite{wang2024instancediffusion} and MS-Diffusion \cite{wang2024ms} also incorporate learnable blocks to handle per-instance conditioning. These methods generally deliver strong performance in precise regional control but are \textit{limited to specific base models} due to the introduction of extra trainable components. On the other hand, tuning-free methods, such as Multidiffusion \cite{bar2023multidiffusion}, RPG \cite{yang2024mastering}, and Omost \cite{omost}, operate on the denoised latent space or attention score map with a mask for each region. They frequently employ a split-and-merge strategy but face challenges in \textit{maintaining precise control as the number of regions increases}.

In this paper, we adopt a tuning-free manner and aim to improve its control strength and coherence when dealing with multiple regions. Specifically, we present \textbf{RAG}, a novel Regional-Aware text-to-image Generation method for precise regional control, which is composed of two sub-tasks, \textbf{Regional Hard Binding} and \textbf{Regional Soft Refinement}. 
First, we implement region-aware hard binding at the beginning of the denoising process to ensure that \textit{each regional prompt is executed accurately}. This step breaks down the input prompt into several regional prompts, each with its respective spatial position, and then merges the individually denoised regional latents into the original image latent.
Second, to \textit{dismiss the visual boundaries and enhance interaction} between adjacent regions, regional soft refinement is applied within the cross-attention layers at the subsequent steps to obtain a regional latent, where $K$ and $V$ are from regional text tokens while $Q$ is from original image latent, followed by a weighted recombination of base image latent and regional latent. 
Furthermore, leveraging robust control and fusion capabilities, our framework supports users to refine specific unsatisfactory regions from the last generation while keeping all other regions intact. 
Both quantitative and qualitative results demonstrate our superior performance over attribute binding and object relationship than previous state-of-the-art tuning-free methods. Our contributions are summarized as follows:

\begin{itemize}
  \item 
    We propose \textbf{RAG}, a tuning-free Regional-Aware text-to-image Generation framework on the top of DiT-based model (FLUX.1-dev), with two novel components, Regional Hard Binding and Regional Soft Refinement, for precise and harmonious regional control.
  \item 
    RAG novelly makes image repainting feasible, allowing users to modify specific unsatisfactory regions in the previous generation while keeping all other regions intact without need for additional inpainting models.
  \item
    Extensive qualitative and quantitative experiments demonstrate that RAG shows superior performance over attribute binding, object relationship and complex composition on T2ICompBench benchmark, in comparison with previous state-of-the-arts.
\end{itemize}
\begin{figure*}[t]
    \centering
    \captionsetup{type=figure}
    \includegraphics[width=\textwidth]{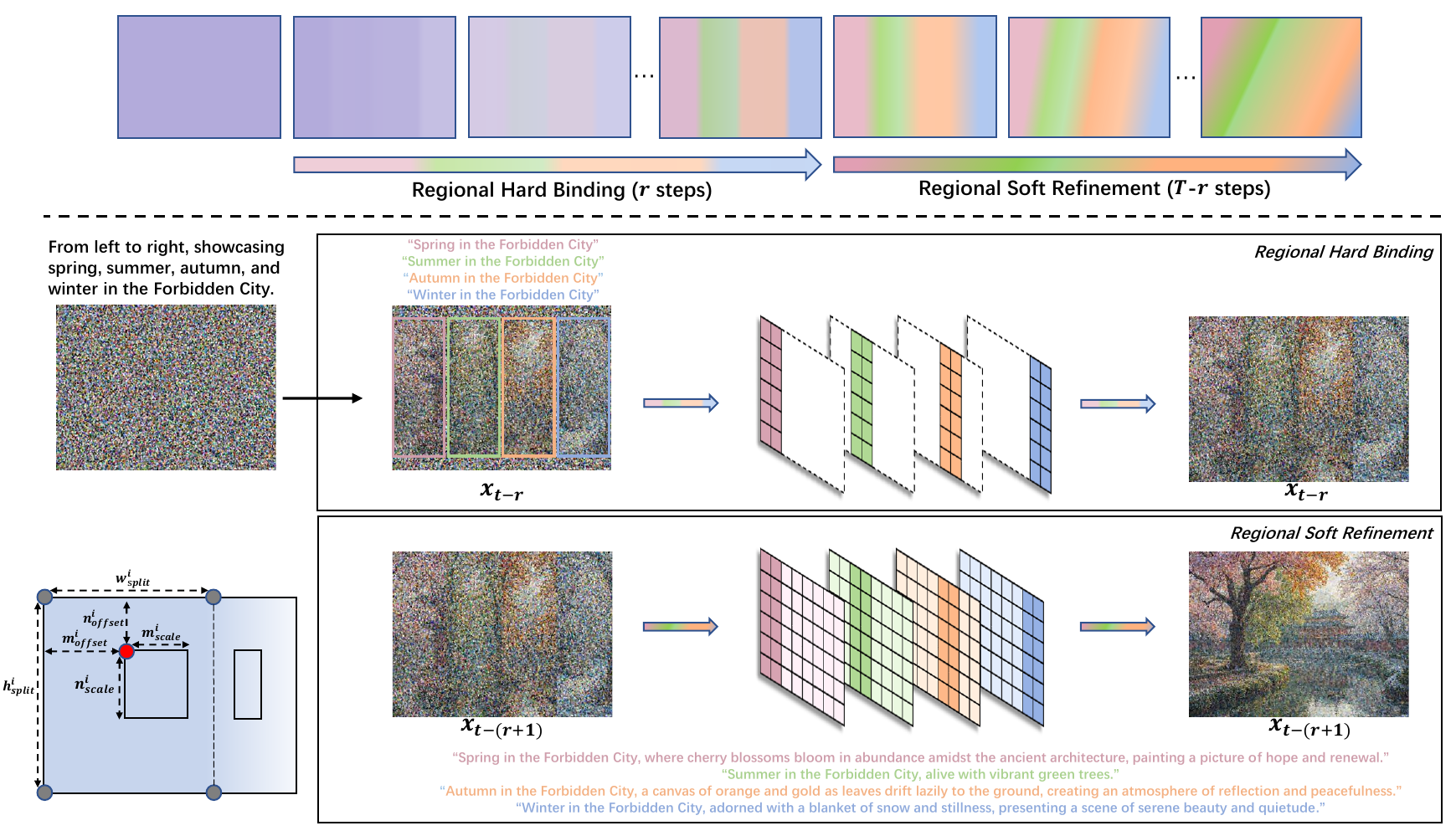}
    \caption{\textbf{The overall framework of RAG}, which divides regional-aware generation into two stages: (1) \textbf{Regional Hard Binding} ensures the proper response of regional prompts by processing each region individually with its fundamental description, and bound at the first $r$ steps to ensure accurate attribute representation and entity localization.
    (2) \textbf{Regional Soft Refinement} improves the harmony of adjacent region via enabling the interaction of regional local conditions with global image latent within the cross-attention layers at the later $T-r$ steps. The lower left corner shows the definition of spatial region in regional hard binding and regional soft refinement. 
    }
    \label{Fig:Overview}
\end{figure*}

\section{Related Work}
\paragraph{Tuning-based Regional Control.}
Conventional text-to-image generation only uses text as conditional input and injects control signals through cross-attention. In order to handle spatial conditions, some works introduce additional training modules, such as ControlNet \cite{zhang2023adding}, to handle new control conditions, including depth maps, sketches, human poses, etc. 
For regional control, additional module is also introduced for training-based methods \cite{zhou2024migc, yang2023reco, feng2024ranni, li2023gligen, wang2024instancediffusion, wang2024ms, xie2023boxdiff, chefer2023attend} to process spatial positions, such as bounding boxes (coordinates) or segmentation masks. 
GLIGEN \cite{li2023gligen} adopts Fourier embedding to encode box coordinates and adds trainable gated self-attention layer at each transformer block to accept new grounding input. InstanceDiffusion \cite{wang2024instancediffusion} further allows diverse ways to specify region positions such as simple single points, scribbles, bounding boxes or segmentation masks. MS-Diffusion \cite{wang2024ms} integrates grounding tokens with its grounding resampler to correlate specific entities and spatial constraints. Other works, such as BoxDiff \cite{xie2023boxdiff} and Attend-and-Excite \cite{chefer2023attend}, apply regional constraints via gradient optimization within the denoising process to ensure all regional tokens are attended.
\vspace{-3mm}

\paragraph{Tuning-Free Regional Control.}
Although tuning-based methods show strong performance, collecting training data is time-consuming and labor-intensive, and they are limited to specific models due to the introduction of additional modules on the top of base models. To address these challenges, model agnostic approaches \cite{kim2023dense,omost,jimenez2023mixture,bar2023multidiffusion,yang2024mastering,feng2024layoutgpt} are proposed. DenseDiffusion \cite{kim2023dense} and Omost \cite{omost} directly adjust attention scores within cross-attention layers to ensure that the activations within masked regions are promoted. Mixture of Diffusers \cite{jimenez2023mixture} and MultiDiffusion \cite{bar2023multidiffusion} denoise different locations separately and then combine the denoised latents using regional masks. 
Recently, RPG \cite{yang2024mastering} introduces complementary regional diffusion through a resize-and-concatenate approach for region-specific compositional generation, where regional denoised latents are resized and merged as a single concatenated latent at each step. However, the control strength of these methods decreases significantly when the number of regions increases. Thus, in our work, we aim to absorbs the advantages of model-agnostic tuning-free methods and improves their control capabilities when dealing with multiple regions.
\section{Method}
\subsection{Preliminaries}

\paragraph{Diffusion Transformer (DiT).} DiT \cite{peebles2023scalable} is a novel architecture integrating transformer as the backbone network within Latent Diffusion Model (LDM), and has become a dominated choice in recent text-to-image generation models like Stable Diffusion 3/3.5 \cite{esser2024scaling} and Flux \cite{flux}. By leveraging transformer, DiT efficiently captures complex dependencies in data, resulting in high-quality image generation. Consistent with the design philosophy of LDM, DiT also operates directly in the latent space, allowing the model to generate high-fidelity images that adhere to specified conditions while reducing computational overhead. 

\vspace{-4mm}
\paragraph{Attention Mechanism.}
The attention mechanism is a crucial component in DiT, enabling effective interactions between the diffusion network and additional control signals such as text or image. During the diffusion process, the attention mechanism incrementally captures feature representations in the latent space, facilitating efficient denoising at each step while preserving global consistency and detail accuracy in the generated results. The attention weights are calculated as follows:
\begin{equation}
    \operatorname{Attention}(Q, K, V)=\operatorname{softmax}\left(\frac{Q K^{T}}{\sqrt{d}}\right) \cdot V,
\end{equation}
where the query matrix $Q$, key matrix $K$, and value matrix $V$ are derived from the input feature embedding vectors in self-attention layers, while $K$ and $V$ are from conditioned embedding vectors in cross-attention layers.

\subsection{Overview of RAG}
We briefly illustrate the idea of RAG as shown in Figure \ref{Fig:Overview}, which decouples the compositional generation process into the construction of individual regions and detail refinement. The implementation of RAG consists of the following steps: 

\noindent \textbf{(1) Regional Hard Binding} (Sec.~\ref{Regional Hard Binding}): This step involves decomposing the raw and complex input prompt containing multiple objects into a subset of \textit{fundamental descriptions} for each individual region or object, along with their corresponding spatial positions. This process can be accomplished either through a finetuned MLLM as done in \cite{yang2024mastering,omost} or through manual definition. Then, each region is processed individually with its fundamental description and bound only at the early stage of denoising to ensure accurate attribute representation and entity localization.

\noindent \textbf{(2) Regional Soft Refinement} (Sec.~\ref{Regional Soft Refinement}): At this step, \textit{highly descriptive} sub-prompts are generated for each region along with a global prompt.  Similarly, this process can be automated using an MLLM or defined manually, further enriching the definition of each object and promote the fusion between adjacent regions. Instead of manipulating the image latent, the refinement step achieves the interaction of regional local conditions and global image latent within the cross-attention layers.

It is worth noting that based on our setting, RAG can novelly support image repainting (Sec.~\ref{Object Repainting}) in a free lunch manner by only re-initialize the initial noise in target areas, thereby enabling accurate modification of previously generated images while maintaining the overall generation quality without the need for an additional inpainting model.

\subsection{Regional Hard Binding}
\label{Regional Hard Binding}
To ensure the proper response of regional prompts and mitigate \textit{object omission} when the number of region or object increases, we apply regional hard binding in the early steps of denoising process as illustrated in Figure \ref{Fig:Overview}, which involves separately denoising the regions with their short fundamental descriptions and then binding local regional latents into global latent.

Specifically, we first decompose the long input prompt $P$ containing multiple objects into a set of fundamental descriptions $\hat{p}^{i}$ with their position sets $m^{i} = \{m_{offset}^{i}, m_{scale}^{i}\}$ and $n^{i} = \{n_{offset}^{i}, n_{scale}^{i}\}$ by MLLM or manually. Subsequently, we perform text encoding on $P$ and $\hat{p}^{i}$ to obtain $y$ and $\hat{y}^{i}$. The the individual latent $\hat{x}^{i}$ utilizes $\hat{y}_{i}$ as a text condition, while origin latent $x$ takes the long input prompt $P$ as condition. The formulaic process is as follows:
\begin{equation}
    {x}_{t-r}={x}_{t-r+1}-\epsilon_{\theta}\left({x}_{t-r+1}, {y}\right)
\end{equation}
\begin{equation}
    \hat{x}^{i}_{t-r}=\hat{x}^{i}_{t-r+1}-\epsilon_{\theta}\left(\hat{x}^{i}_{t-r+1}, \hat{y}^{i}\right),
\end{equation}
where $i\in[1,k]$, $k$ is the number of regions. $\epsilon_\theta$ is the noise predicted.
For each denoising step, we bind $\hat{x}^{i}_{t-r}$ to the latent space in the rectangular area given by $m_{i}$,$n_{i}$ as follows:
\begin{equation}
    x_{t-r} = \operatorname{Replace}(x_{t-r}, \hat{x}^{i}_{t-r}, m^{i}, n^{i}),
\end{equation}
where $\operatorname{Replace(\cdot)}$ represents the process of pasting the individual latents back to the corresponding area in origin latent. The binding is only executed in the early steps within denoising process. We find that a few steps of binding is sufficient for regional completeness, whereas full-steps binding results in either clear visual boundaries adjacent regions or poor interactivity.

\begin{table*}[!ht]
\centering
\begin{tabular}{ccccccc}
\hline
\multirow{2}{*}{Model}           & \multicolumn{3}{c}{Attribute Binding} & \multicolumn{2}{c}{Object Relationship} & \multirow{2}{*}{Complex $\uparrow$} \\ \cline{2-6}
                                 & Color $\uparrow$      & Shape $\uparrow$      & Texture $\uparrow$     & Spatial $\uparrow$          & Non-Spatial $\uparrow$          &                          \\ \hline
\multicolumn{1}{c|}{Stable v1.4 \cite{rombach2022high}\hspace{0.2em}~\pub{CVPR 2022}} & 0.3765     & 0.3576     & 0.4156      & 0.1246           & 0.3079               & 0.3080                   \\
\multicolumn{1}{c|}{Composable v2 \cite{liu2022compositional}\hspace{0.2em}~\pub{ECCV 2022}} & 0.4063     & 0.3299     & 0.3645      & 0.0800           & 0.2980               & 0.2898                   \\
\multicolumn{1}{c|}{Structured v2 \cite{feng2022training}\hspace{0.2em}~\pub{ICLR 2023}} & 0.4990     & 0.4218     & 0.4900      & 0.1386           & 0.3111               & 0.3355                   \\
\multicolumn{1}{c|}{Stable v2 \cite{rombach2022high}\hspace{0.2em}~\pub{CVPR 2022}}   & 0.5065     & 0.4221     & 0.4922      & 0.1342           & 0.3127               & 0.3386                   \\
\multicolumn{1}{c|}{Stable XL \cite{betker2023improving}\hspace{0.2em}~\pub{2022}}   & 0.5879     & 0.4687     & 0.5299      & 0.2133           & 0.3119               & 0.3237                   \\
\multicolumn{1}{c|}{Attn-Exct v2 \cite{chefer2023attend}\hspace{0.2em}~\pub{TOG 2023}}   & 0.6400     & 0.4517     & 0.5963      & 0.1455           & 0.3109               & 0.3401 \\
\multicolumn{1}{c|}{GORS \cite{huang2023t2i}\hspace{0.2em}~\pub{Neurips 2023}}   & 0.6603     & 0.4785     & 0.6287      & 0.1815           & 0.3193               & 0.3328\\
\multicolumn{1}{c|}{Pixart-$\alpha$-ft \cite{chen2023pixart}\hspace{0.2em}~\pub{ICLR 2024}}      & 0.6690     & 0.4927     & 0.6477      & 0.2064           & \underline{0.3197}               & 0.3433                   \\
\multicolumn{1}{c|}{RPG* \cite{yang2024mastering}\hspace{0.2em}~\pub{ICML 2024}}         & 0.7476     & \underline{0.5640}     & \underline{0.6724}      & \underline{0.4017}           & 0.3032               & \underline{0.3702}                   \\
\multicolumn{1}{c|}{Flux.1-dev* \cite{flux}\hspace{0.2em}~\pub{2024}}  & \underline{0.7680}     & 0.5078     & 0.6195      & 0.2606           & 0.3078               & 0.3650                   \\ \hline
\multicolumn{1}{c|}{Ours}        & \textbf{0.8039}     & \textbf{0.6016}     & \textbf{0.7085}      & \textbf{0.5193}           & \textbf{0.3263}               & \textbf{0.4377}                   \\ \hline
\end{tabular}
\caption{Comparison of alignment evaluation on T2ICompBench~\cite{huang2023t2i}. 
	The best results are highlighted in $\mathbf{bold}$, second-best in \underline{underline}. 
	The basic data is sourced from~\cite{huang2023t2i}.
	* indicates results we reproduced using the official open-source codes and configurations.}
\label{Table:Quantitative Comparison}
\end{table*}

\subsection{Regional Soft Refinement}
\label{Regional Soft Refinement}

Images generated by direct regional hard binding allow for precise control over positioning, effectively preventing object omission. However, the rendering of attributes tends to be relatively coarse, and there is a tendency for \textit{noticeable boundaries between adjacent regions}. Therefore, we apply regional soft refinement in the later steps of the denoising process to improve the harmony of adjacent regions.

Specifically, similar to the previous binding step, we break down the original long prompt $P$, but instead of breaking it down into short fundamental descriptions, we break it down into highly descriptive sub-prompts $\tilde{p}^{i}$, with a set of global regions $o^{i} = \{h_{split}^{i}, w_{split}^{i}\}$, and obtain the corresponding representation $\tilde{y}^{i}$ through the text encoder. Then, the text condition $\tilde{y}^{i}$ is 
projected into $K^{i}$ and $V^{i}$, while $Q^{i}$ is derived from current image latent $x_{t-(r+1)}$, within cross-attention layers:
\begin{equation}
    \begin{split}
        Q^{i}=\ell_{Q}\left(x_{t-(r+1)}\right), K^{i}=\ell_{K}(\tilde{y}^{i}) , V^{i}=&\ell_{V}(\tilde{y}^{i})  \\         
        \mathbf{x}_{t-(r+1)}^{i}=\operatorname{Softmax}\left(\frac{Q^{i} K^{{i}^{T}}}{\sqrt{d}}\right) & V^{i},
    \end{split}
\end{equation}
where $\ell_{Q}$, $\ell_{k}$, $\ell_{V}$ are linear projections, $\mathbf{x}_{t-(r+1)}^{i}$ is globally-informed latent.
We crop $\mathbf{x}_{t-(r+1)}^{i}$ to the region corresponding to different objects to obtain  $\tilde{\mathbf{x}}_{t-(r+1)}^{i}$ with rich attributes generated in the global region:
\begin{equation}
    \tilde{\mathbf{x}}_{t-(r+1)}^{i} = \operatorname{Crop}\left(\mathbf{x}_{t-(r+1)}^{i}, o^{i} \right).
\end{equation}

Then, we also execute $\operatorname{Replace(\cdot)}$ to splice these regions:
\begin{equation}
    x^{r}_{t-(r+1)}=\operatorname{Replace}\left(x_{t-(r+1)}, \tilde{x}_{t-(r+1)}^{i}, o^{i} \right).
\end{equation}

To further improve the alignment between each region with the original prompt and enhance adjacent interactions, we perform a linear weighted recombination of $x^{r}_{t-(r+1)}$ and $x_{t-(r+1)}$ as follows:
\begin{equation}
    x_{t-(r+1)} = x_{t-(r+1)}*(1-\delta)_ + x^{r}_{t-(r+1)}*\delta,
\end{equation}
where $\delta$ controls the fusion strength of image latent $x_{t-(r+1)}$ and region-aware global latent $x^{r}_{t-(r+1)}$.

\begin{figure}
    \centering
    \includegraphics[width=\linewidth]{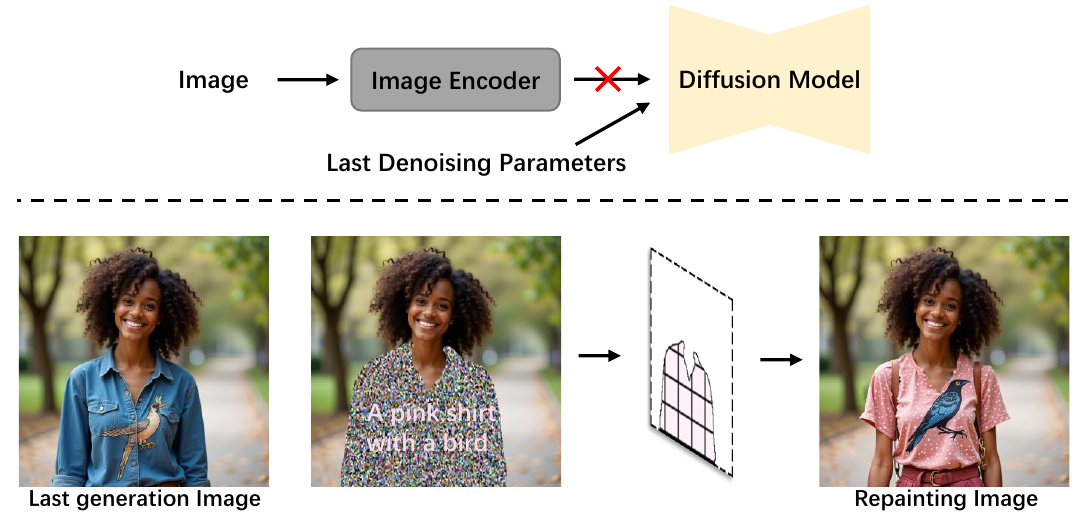}
    \caption{\textbf{Illustration of Re-painting}. Different from regular image-to-image inpainting, repainting inherits from last generation with only the target area re-initialized (upper). Given the parameters in previous generation, users are allowed to specify a target area with a new prompt and repaint the image, without relying on additional inpainting models (bottom).}
    \label{Fig:repainting}
\end{figure}

\subsection{Image Repainting}
\label{Object Repainting}

Based on our task setting, where a long complex prompt containing multiple objects is decoupled into several regional sub-prompts for individual processing, a natural question is raised: \textit{Can this approach be used to repaint specific areas?} 
By leveraging the robust control and fusion capabilities of regional hard binding and regional soft refinement, we can reinitialize noise in a specific region requiring modification, enabling repainting of a region without affecting the overall layout or attributes of other areas.
Different from typical post-processing inpainting task that usually requires additional inpainting models, repainting re-generates with the same parameters from the last generation, with only the prompts in the target area modified.

As shown in Figure \ref{Fig:repainting}, users are allowed to modify the description of a specific region within $P$, $\hat{p}^{i}$, $\tilde{p}^{i}$ to obtain new edited prompts $P_{e}$, $\hat{p}^{i}_{e}$, $\tilde{p}^{i}_{e}$. Subsequently, we encode them as text features $y_{e}$, $\hat{y}^{i}_{e}$, $\tilde{y}^{i}_{e}$. Given a total denoising step $T$, the initial noisy latent $x^{'}_{T}$ is inherited from the previous generation $x_{T}$, only the target area indicated by $mask$ is re-initialized.
\begin{equation}
    x^{'}_{T} =  \operatorname{init}(x_{T}, mask).
\end{equation}

To ensure the other regions intact, we simultaneously perform denoising on $x_{T}$ and $x^{'}_{T}$. At each timestep $t$, we replace the corresponding portions of $x_t$ with the masked areas of the edited region from $x^{'}_{t}$ for repainting:
\begin{equation}
    x_t = \operatorname{Repaint}(x_t, x^{'}_{t}, mask).
\end{equation}

Through the regional complementarity enabled by regional soft refinement, the repainting region can be seamlessly integrated with the surrounding areas of other regions. This approach not only \textit{makes image repainting feasible}, but also \textit{ensures the coherence and consistency of the overall scene}.
\begin{figure*}[t]
    \centering
    \includegraphics[width=\textwidth]{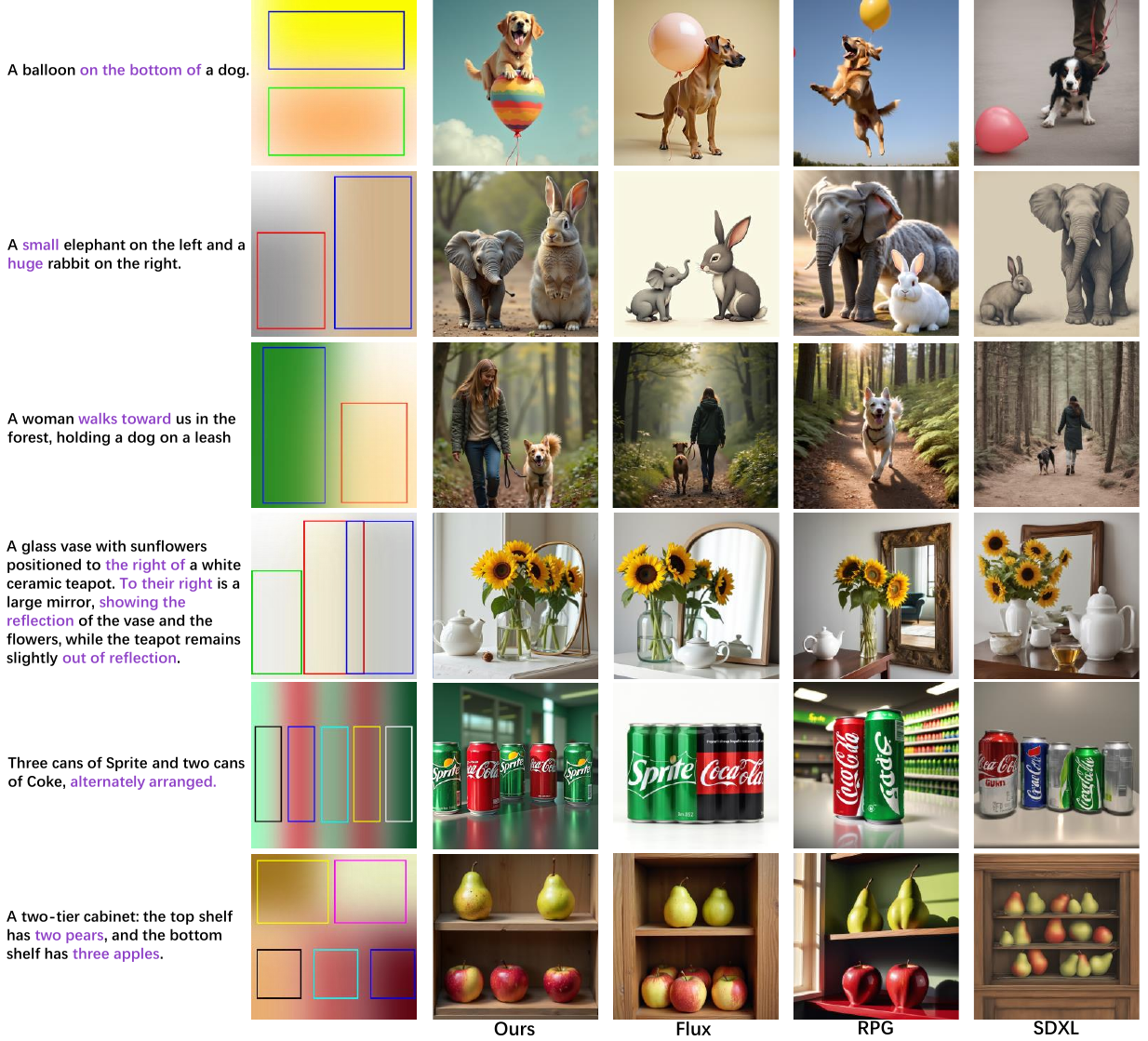}
    \caption{\textbf{Qualitative comparisons on compositional text-to-image generation}. From top to bottom, we show 6 examples of different prompts and regions. Compared with previous methods, we demonstrate excellent regional control capabilities.}
    \label{Fig:Qualitative_Comparison}
\end{figure*}

\section{Experiments}
\label{sec:Experiments}
\subsection{Experiment Setting}
\paragraph{Implementation Details.} 
We implement our approach on the top of Flux.1-dev \cite{flux} for its superior performance.
The inference process is set to 20 steps, with guidance scale of 3.5. 
For large-scale quantitative evaluations, we employ Chain-of-Thought (CoT) template from \cite{yang2024mastering} and leverage GPT-4 to automatically decouple multi-object scenes. 
All experiments are conducted on a single A6000 GPU.

\vspace{-3mm}
\paragraph{Compared Methods.} To comprehensively evaluate the generation quality, we compare our RAG with several state-of-the-art text-to-image approaches, including: Stable v1.4 \cite{rombach2022high}, Composable v2 \cite{liu2022compositional}, Structured v2 \cite{feng2022training}, Stable v2 \cite{rombach2022high}, Stable XL \cite{betker2023improving}, Attn-Exct v2 \cite{chefer2023attend}, {GORS \cite{huang2023t2i}, Pixart-$\alpha$-ft \cite{chen2023pixart}, RPG \cite{yang2024mastering}, and Flux.1-dev \cite{flux}.

\begin{figure*}[t]
    \centering
    \captionsetup{type=figure}
    \includegraphics[width=\textwidth]{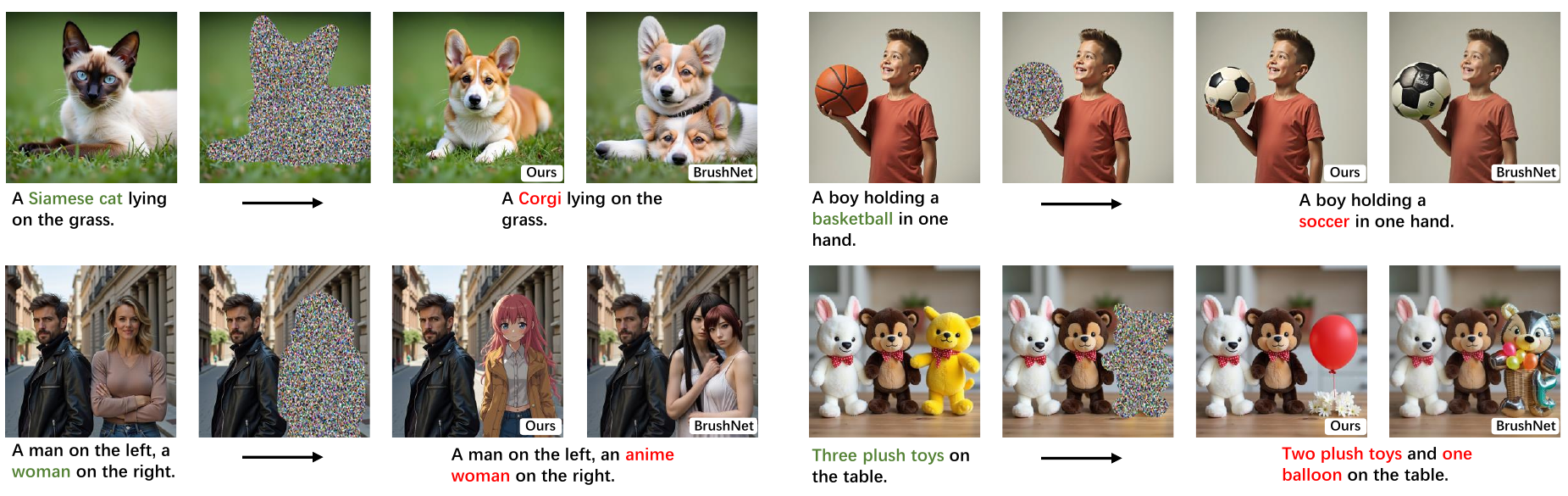}
    \caption{\textbf{Qualitative comparisons on image repainting} between our RAG and the state-of-the-art inpainting model BrushNet. Our results are more region-aware with harmonious effect with the surrounding, revealing diverse potential for applications.}
    \label{Fig:Qualitative Comparison-repainting}
\end{figure*}

\subsection{Main Results}
\paragraph{Quantitative Comparison.} 
Table~\ref{Table:Quantitative Comparison} presents our alignment evaluation on the T2ICompBench benchmark, comparing RAG with other state-of-the-art methods.
Our approach outperforms competitors in key aspects such as attribute binding, object relationships, and complex composition. Notably, RAG achieves a 29\% improvement over RPG in handling prompts with spatial relationships, underscoring its effectiveness in spatially accurate generation.
Quantitative results demonstrate our superior performance in handling complex multi-region prompts.

\paragraph{Qualitative Comparison.} 
Figure~\ref{Fig:Qualitative_Comparison} illustrates visual comparisons, highlighting our superiority in complex multi-region generation. While RPG~\cite{yang2024mastering} also excels at regional control, its lack of precise positional control may lead to object omission or fusion. In contrast, our method enable accurate multi-region control via hard binding and soft refinement mechanisms, faithfully conveying details on position, quantity, and attributes based on the input text. Figure~\ref{Fig:Qualitative Comparison-repainting} shows the comparison with the state-of-the-art inpainting model BrushNet \cite{ju2024brushnet}, showcasing our strength in repainting \textit{both single and contacted regions without conflict}. Even when the repainted object differs greatly in style or shape, our method successfully generates the intended target while maintaining the layout and regional attributes, enhancing the flexibility and control of the generation process.

\begin{table}[h]
\centering
\begin{tabular}{ccc}
\hline
Methods                   & Aesthetic $\uparrow$ & Alignment $\uparrow$ \\ \hline
\multicolumn{1}{c|}{RPG \cite{yang2024mastering}\hspace{0.2em}~\pub{ICML 2024}}  & 16.1\%     & 14.8\%     \\
\multicolumn{1}{c|}{Stable v3 \cite{esser2024scaling}\hspace{0.2em}~\pub{ICML 2024}}  & 13.2\%    & 13.4\%     \\
\multicolumn{1}{c|}{Flux.1-dev \cite{flux}\hspace{0.2em}~\pub{2024}} & 18.8\%     & 17.8\%    \\ \hline
\multicolumn{1}{c|}{Ours} & \textbf{51.9\%}     & \textbf{54.0\%}     \\ \hline
\end{tabular}
\caption{User study on aesthetics and text-image alignment. Our approach shows a significant improvement.}
\label{user study}
\end{table}

\vspace{-3mm}
\paragraph{User Study.} 
We randomly sampled 109 prompts from the T2ICompBench test set to evaluate the aesthetic quality and text-image consistency (including attributes and spatial positions) of the generated images. During the evaluation, we invited 24 users to compare the generation results of RPG \cite{yang2024mastering} and Stable v3 \cite{esser2024scaling}, Flux.1-dev \cite{flux}, RAG, selecting the most suitable generation results for each prompt. Through this blind test, we obtained 2,616 user votes. The results are shown in Table \ref{user study}. RAG is far ahead in terms of both aesthetics and alignment.

\vspace{-3mm}
\subsection{Ablation Study}
\paragraph{Effectiveness of Hard Binding \& Soft Refinement.} 
The proposed RAG comprises two key components: hard binding and soft refinement. 
We conducted ablation experiments on these two parts, with visual results shown in Figure~\ref{Fig:all_ablation}.  
Results reveal that hard binding achieves precise object positioning but yields coarse attribute rendering and limited integration among objects.
With soft refinement, object attributes are richly detailed and relationships are harmonious, though precise positioning may be compromised, and object omission may sometimes occur.

\begin{figure}
    \centering
    \includegraphics[width=\linewidth]{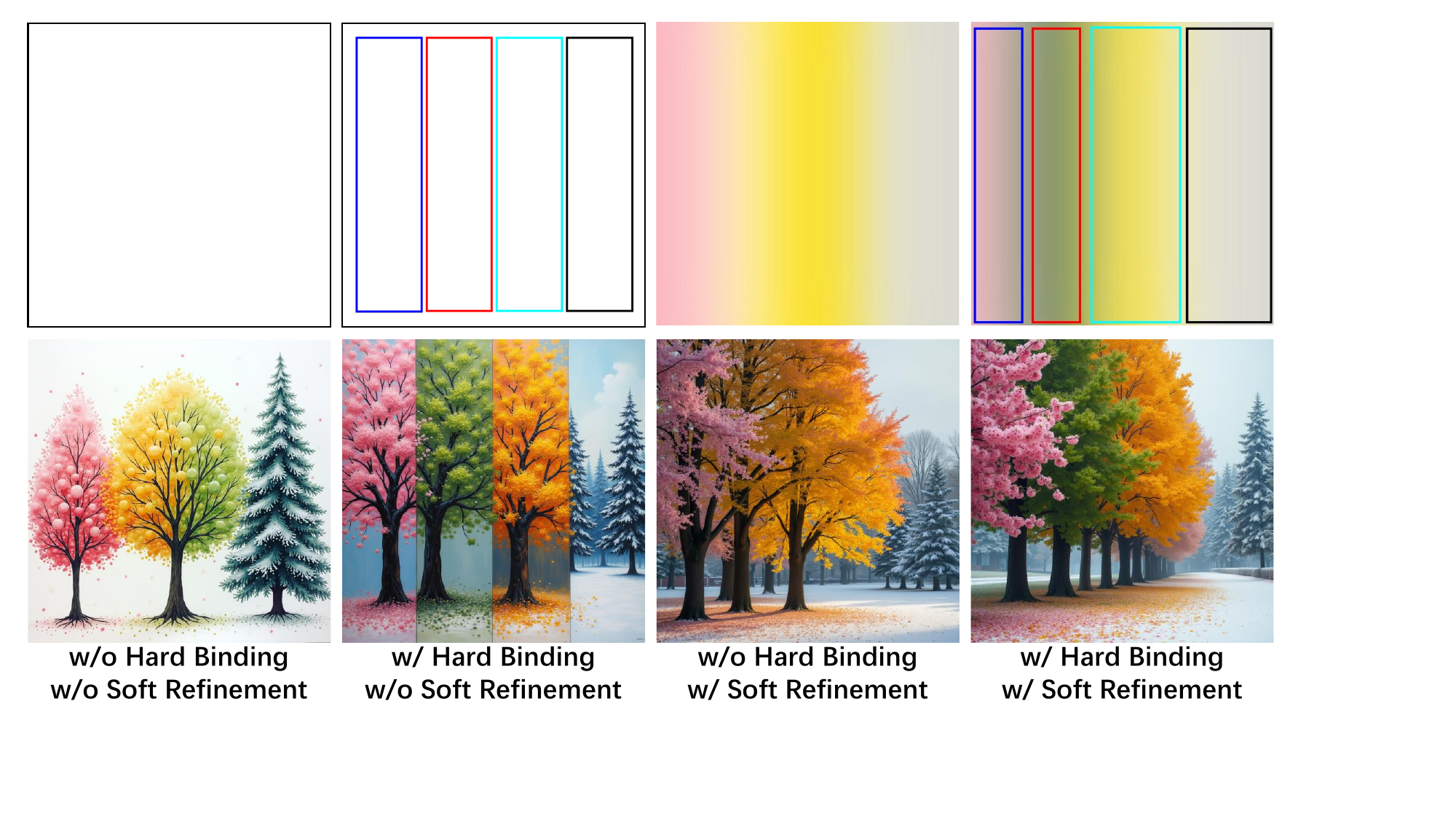}
    \caption{\textbf{Qualitative analysis of Hard Binding and Soft Refinement}. The former ensures the proper responses of each region, while the latter enhances the coherence among regions.}
    \label{Fig:all_ablation}
\end{figure}

\begin{figure*}[t]
    \centering
    \captionsetup{type=figure}
    \includegraphics[width=\textwidth]{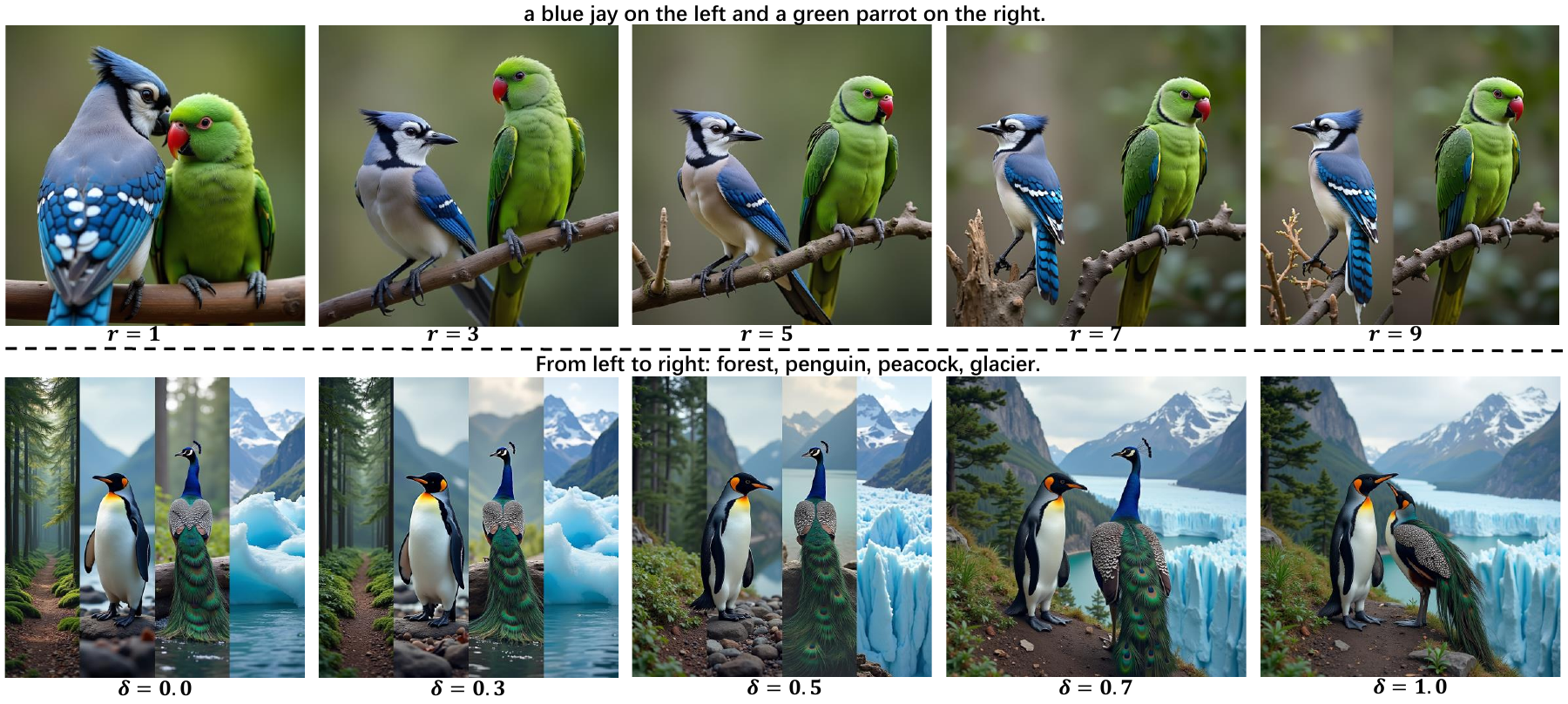}
    \caption{Qualitative analysis of hard binding steps $r$ and soft refinement strength $\delta$. A few steps of binding is sufficient for regional completeness, and appropriate soft refinement intensity leads to improved regional coherence.}
    \label{Fig:r_delta_ablation}
\end{figure*}

\begin{figure}
    \centering
    \includegraphics[width=\linewidth]{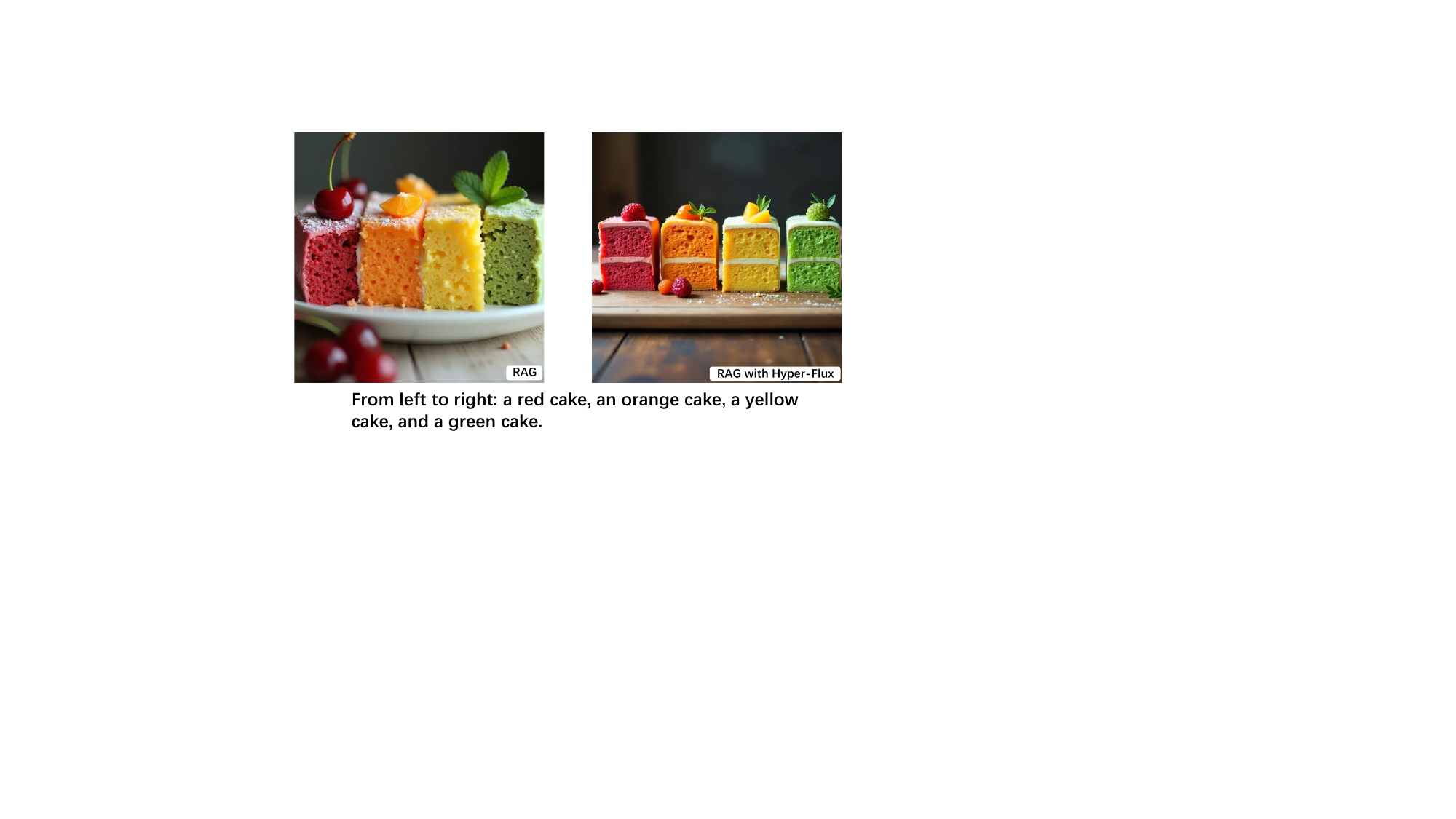}
    \caption{RAG, accelerated by Hyper-Flux, is 2.5$\times$ faster than the original version, maintaining visual quality.}
    \label{Fig:speed and visiual}
\end{figure}

\vspace{-3mm}
\paragraph{Effectiveness of Parameter $r$.} 
We introduced a parameter $r$ to control the frequency of hard binding applications. 
As shown in Figure \ref{Fig:r_delta_ablation}, excessive hard binding limits the opportunity for subsequent soft refinement, weakening its impact.
This may diminish interactions between adjacent objects, potentially causing noticeable boundaries to reappear. 
By adjusting $r$, we aim to balance precise object placement with smooth integration, optimizing both the structure and cohesion of the generated image.

\vspace{-3mm}
\paragraph{Effectiveness of Coefficient $\delta$.} 

We introduced the coefficient $\delta$ to modulate the intensity of regional soft refinement. As illustrated in  Figure~\ref{Fig:r_delta_ablation}, excessively low or high $\delta$ values may result in noticeable partitioning or slight misalignment between text and image. Setting $\delta$ to an optimal level enhances the fusion of image regions, yielding a more coherent and natural overall output.

\vspace{-3mm}
\paragraph{Analysis of Inference Time.} 
In RAG, the decoupled multi-region generation operation processes each region independently, leading to increased inference time as the number of regions grows. However, because RAG does not modify the underlying architecture of the original model, existing acceleration frameworks, such as Hyper-Flux \cite{ren2024hyper}, remain compatible with RAG. As shown in Figure \ref{Fig:speed and visiual}, combining RAG with Hyper-Flux results in inference being 2.5-fold faster while preserving the stability of accuracy and detail control in regional generation. This demonstrates that, although inference speed presents a challenge, with appropriate acceleration strategies, RAG can efficiently generate multi-region images with high quality and maintain control over interactions between regions.
\section{Conclusion}
\label{sec:Conclusion}

In this paper, we introduce RAG, a novel tuning-free Region-Aware text-to-image Generation method designed to address challenges posed by regional prompts. RAG operates through two key stages. First, regional hard binding independently constructs the content of each region.
Then, regional soft refinement enhances interactions between adjacent regions and improves attribute generation.
Furthermore, RAG bypasses the need for external inpainting models, enabling direct image repainting to modify unsatisfactory regions from last generations. 
Extensive experimental results demonstrate the superiority of RAG on compositional generation compared to prior tuning-free methods.

\vspace{-3mm}
\paragraph{Limitation and Future Work.}
\label{sec:Limitation}
RAG offers precise regional control and flexible image repainting but has limitations. Its multi-region decoupling increases inference time as region count grows. Future work will focus on improving RAG’s inference efficiency and integrating with other diffusion models for enhanced scalability and performance.

{
    \small
    \bibliographystyle{ieeenat_fullname}
    \bibliography{main}

\begin{thebibliography}{44}
\providecommand{\natexlab}[1]{#1}
\providecommand{\url}[1]{\texttt{#1}}
\expandafter\ifx\csname urlstyle\endcsname\relax
  \providecommand{\doi}[1]{doi: #1}\else
  \providecommand{\doi}{doi: \begingroup \urlstyle{rm}\Url}\fi

\bibitem[Bar-Tal et~al.(2023)Bar-Tal, Yariv, Lipman, and Dekel]{bar2023multidiffusion}
Omer Bar-Tal, Lior Yariv, Yaron Lipman, and Tali Dekel.
\newblock Multidiffusion: Fusing diffusion paths for controlled image generation.
\newblock 2023.

\bibitem[Betker et~al.(2023)Betker, Goh, Jing, Brooks, Wang, Li, Ouyang, Zhuang, Lee, Guo, et~al.]{betker2023improving}
James Betker, Gabriel Goh, Li Jing, Tim Brooks, Jianfeng Wang, Linjie Li, Long Ouyang, Juntang Zhuang, Joyce Lee, Yufei Guo, et~al.
\newblock Improving image generation with better captions.
\newblock \emph{Computer Science. https://cdn. openai. com/papers/dall-e-3. pdf}, 2\penalty0 (3):\penalty0 8, 2023.

\bibitem[BlackForest(2024)]{flux}
BlackForest.
\newblock Black forest labs; frontier ai lab, 2024.

\bibitem[Chefer et~al.(2023)Chefer, Alaluf, Vinker, Wolf, and Cohen-Or]{chefer2023attend}
Hila Chefer, Yuval Alaluf, Yael Vinker, Lior Wolf, and Daniel Cohen-Or.
\newblock Attend-and-excite: Attention-based semantic guidance for text-to-image diffusion models.
\newblock \emph{ACM Transactions on Graphics (TOG)}, 42\penalty0 (4):\penalty0 1--10, 2023.

\bibitem[Chen et~al.(2023)Chen, Yu, Ge, Yao, Xie, Wu, Wang, Kwok, Luo, Lu, et~al.]{chen2023pixart}
Junsong Chen, Jincheng Yu, Chongjian Ge, Lewei Yao, Enze Xie, Yue Wu, Zhongdao Wang, James Kwok, Ping Luo, Huchuan Lu, et~al.
\newblock Pixart-$\alpha$: Fast training of diffusion transformer for photorealistic text-to-image synthesis.
\newblock \emph{arXiv preprint arXiv:2310.00426}, 2023.

\bibitem[Du et~al.(2021)Du, Qian, Liu, Ding, Qiu, Yang, and Tang]{du2021glm}
Zhengxiao Du, Yujie Qian, Xiao Liu, Ming Ding, Jiezhong Qiu, Zhilin Yang, and Jie Tang.
\newblock Glm: General language model pretraining with autoregressive blank infilling.
\newblock \emph{arXiv preprint arXiv:2103.10360}, 2021.

\bibitem[Dubey et~al.(2024)Dubey, Jauhri, Pandey, Kadian, Al-Dahle, Letman, Mathur, Schelten, Yang, Fan, et~al.]{dubey2024llama}
Abhimanyu Dubey, Abhinav Jauhri, Abhinav Pandey, Abhishek Kadian, Ahmad Al-Dahle, Aiesha Letman, Akhil Mathur, Alan Schelten, Amy Yang, Angela Fan, et~al.
\newblock The llama 3 herd of models.
\newblock \emph{arXiv preprint arXiv:2407.21783}, 2024.

\bibitem[Esser et~al.(2024)Esser, Kulal, Blattmann, Entezari, M{\"u}ller, Saini, Levi, Lorenz, Sauer, Boesel, et~al.]{esser2024scaling}
Patrick Esser, Sumith Kulal, Andreas Blattmann, Rahim Entezari, Jonas M{\"u}ller, Harry Saini, Yam Levi, Dominik Lorenz, Axel Sauer, Frederic Boesel, et~al.
\newblock Scaling rectified flow transformers for high-resolution image synthesis.
\newblock In \emph{Forty-first International Conference on Machine Learning}, 2024.

\bibitem[Feng et~al.(2022)Feng, He, Fu, Jampani, Akula, Narayana, Basu, Wang, and Wang]{feng2022training}
Weixi Feng, Xuehai He, Tsu-Jui Fu, Varun Jampani, Arjun Akula, Pradyumna Narayana, Sugato Basu, Xin~Eric Wang, and William~Yang Wang.
\newblock Training-free structured diffusion guidance for compositional text-to-image synthesis.
\newblock \emph{arXiv preprint arXiv:2212.05032}, 2022.

\bibitem[Feng et~al.(2024{\natexlab{a}})Feng, Zhu, Fu, Jampani, Akula, He, Basu, Wang, and Wang]{feng2024layoutgpt}
Weixi Feng, Wanrong Zhu, Tsu-jui Fu, Varun Jampani, Arjun Akula, Xuehai He, Sugato Basu, Xin~Eric Wang, and William~Yang Wang.
\newblock Layoutgpt: Compositional visual planning and generation with large language models.
\newblock \emph{Advances in Neural Information Processing Systems}, 36, 2024{\natexlab{a}}.

\bibitem[Feng et~al.(2024{\natexlab{b}})Feng, Gong, Chen, Shen, Liu, and Zhou]{feng2024ranni}
Yutong Feng, Biao Gong, Di Chen, Yujun Shen, Yu Liu, and Jingren Zhou.
\newblock Ranni: Taming text-to-image diffusion for accurate instruction following.
\newblock In \emph{Proceedings of the IEEE/CVF Conference on Computer Vision and Pattern Recognition}, pages 4744--4753, 2024{\natexlab{b}}.

\bibitem[Ho et~al.(2020)Ho, Jain, and Abbeel]{ho2020denoising}
Jonathan Ho, Ajay Jain, and Pieter Abbeel.
\newblock Denoising diffusion probabilistic models.
\newblock \emph{Advances in neural information processing systems}, 33:\penalty0 6840--6851, 2020.

\bibitem[Huang et~al.(2023)Huang, Sun, Xie, Li, and Liu]{huang2023t2i}
Kaiyi Huang, Kaiyue Sun, Enze Xie, Zhenguo Li, and Xihui Liu.
\newblock T2i-compbench: A comprehensive benchmark for open-world compositional text-to-image generation.
\newblock \emph{Advances in Neural Information Processing Systems}, 36:\penalty0 78723--78747, 2023.

\bibitem[Jim{\'e}nez(2023)]{jimenez2023mixture}
{\'A}lvaro~Barbero Jim{\'e}nez.
\newblock Mixture of diffusers for scene composition and high resolution image generation.
\newblock \emph{arXiv preprint arXiv:2302.02412}, 2023.

\bibitem[Ju et~al.(2024)Ju, Liu, Wang, Bian, Shan, and Xu]{ju2024brushnet}
Xuan Ju, Xian Liu, Xintao Wang, Yuxuan Bian, Ying Shan, and Qiang Xu.
\newblock Brushnet: A plug-and-play image inpainting model with decomposed dual-branch diffusion.
\newblock \emph{arXiv preprint arXiv:2403.06976}, 2024.

\bibitem[Kim et~al.(2023)Kim, Lee, Kim, Ha, and Zhu]{kim2023dense}
Yunji Kim, Jiyoung Lee, Jin-Hwa Kim, Jung-Woo Ha, and Jun-Yan Zhu.
\newblock Dense text-to-image generation with attention modulation.
\newblock In \emph{Proceedings of the IEEE/CVF International Conference on Computer Vision}, pages 7701--7711, 2023.

\bibitem[Li et~al.(2023)Li, Liu, Wu, Mu, Yang, Gao, Li, and Lee]{li2023gligen}
Yuheng Li, Haotian Liu, Qingyang Wu, Fangzhou Mu, Jianwei Yang, Jianfeng Gao, Chunyuan Li, and Yong~Jae Lee.
\newblock Gligen: Open-set grounded text-to-image generation.
\newblock In \emph{Proceedings of the IEEE/CVF Conference on Computer Vision and Pattern Recognition}, pages 22511--22521, 2023.

\bibitem[Liu et~al.(2024)Liu, Akhgari, Visheratin, Kamko, Xu, Shrirao, Souza, Doshi, and Li]{liu2024playground}
Bingchen Liu, Ehsan Akhgari, Alexander Visheratin, Aleks Kamko, Linmiao Xu, Shivam Shrirao, Joao Souza, Suhail Doshi, and Daiqing Li.
\newblock Playground v3: Improving text-to-image alignment with deep-fusion large language models.
\newblock \emph{arXiv preprint arXiv:2409.10695}, 2024.

\bibitem[Liu et~al.(2022)Liu, Li, Du, Torralba, and Tenenbaum]{liu2022compositional}
Nan Liu, Shuang Li, Yilun Du, Antonio Torralba, and Joshua~B Tenenbaum.
\newblock Compositional visual generation with composable diffusion models.
\newblock In \emph{European Conference on Computer Vision}, pages 423--439. Springer, 2022.

\bibitem[Omost-Team(2024)]{omost}
Omost-Team.
\newblock Omost github page, 2024.

\bibitem[Peebles and Xie(2023)]{peebles2023scalable}
William Peebles and Saining Xie.
\newblock Scalable diffusion models with transformers.
\newblock In \emph{Proceedings of the IEEE/CVF International Conference on Computer Vision}, pages 4195--4205, 2023.

\bibitem[Podell et~al.(2023)Podell, English, Lacey, Blattmann, Dockhorn, M{\"u}ller, Penna, and Rombach]{podell2023sdxl}
Dustin Podell, Zion English, Kyle Lacey, Andreas Blattmann, Tim Dockhorn, Jonas M{\"u}ller, Joe Penna, and Robin Rombach.
\newblock Sdxl: Improving latent diffusion models for high-resolution image synthesis.
\newblock \emph{arXiv preprint arXiv:2307.01952}, 2023.

\bibitem[Raffel et~al.(2020)Raffel, Shazeer, Roberts, Lee, Narang, Matena, Zhou, Li, and Liu]{raffel2020exploring}
Colin Raffel, Noam Shazeer, Adam Roberts, Katherine Lee, Sharan Narang, Michael Matena, Yanqi Zhou, Wei Li, and Peter~J Liu.
\newblock Exploring the limits of transfer learning with a unified text-to-text transformer.
\newblock \emph{Journal of machine learning research}, 21\penalty0 (140):\penalty0 1--67, 2020.

\bibitem[Ramesh et~al.(2022)Ramesh, Dhariwal, Nichol, Chu, and Chen]{ramesh2022hierarchical}
Aditya Ramesh, Prafulla Dhariwal, Alex Nichol, Casey Chu, and Mark Chen.
\newblock Hierarchical text-conditional image generation with clip latents.
\newblock \emph{arXiv preprint arXiv:2204.06125}, 1\penalty0 (2):\penalty0 3, 2022.

\bibitem[Ren et~al.(2024)Ren, Xia, Lu, Zhang, Wu, Xie, Wang, and Xiao]{ren2024hyper}
Yuxi Ren, Xin Xia, Yanzuo Lu, Jiacheng Zhang, Jie Wu, Pan Xie, Xing Wang, and Xuefeng Xiao.
\newblock Hyper-sd: Trajectory segmented consistency model for efficient image synthesis.
\newblock \emph{arXiv preprint arXiv:2404.13686}, 2024.

\bibitem[Rombach et~al.(2022)Rombach, Blattmann, Lorenz, Esser, and Ommer]{rombach2022high}
Robin Rombach, Andreas Blattmann, Dominik Lorenz, Patrick Esser, and Bj{\"o}rn Ommer.
\newblock High-resolution image synthesis with latent diffusion models.
\newblock In \emph{Proceedings of the IEEE/CVF conference on computer vision and pattern recognition}, pages 10684--10695, 2022.

\bibitem[Ronneberger et~al.(2015)Ronneberger, Fischer, and Brox]{ronneberger2015u}
Olaf Ronneberger, Philipp Fischer, and Thomas Brox.
\newblock U-net: Convolutional networks for biomedical image segmentation.
\newblock In \emph{Medical image computing and computer-assisted intervention--MICCAI 2015: 18th international conference, Munich, Germany, October 5-9, 2015, proceedings, part III 18}, pages 234--241. Springer, 2015.

\bibitem[Saharia et~al.(2022)Saharia, Chan, Saxena, Li, Whang, Denton, Ghasemipour, Gontijo~Lopes, Karagol~Ayan, Salimans, et~al.]{saharia2022photorealistic}
Chitwan Saharia, William Chan, Saurabh Saxena, Lala Li, Jay Whang, Emily~L Denton, Kamyar Ghasemipour, Raphael Gontijo~Lopes, Burcu Karagol~Ayan, Tim Salimans, et~al.
\newblock Photorealistic text-to-image diffusion models with deep language understanding.
\newblock \emph{Advances in neural information processing systems}, 35:\penalty0 36479--36494, 2022.

\bibitem[Sohl-Dickstein et~al.(2015)Sohl-Dickstein, Weiss, Maheswaranathan, and Ganguli]{sohl2015deep}
Jascha Sohl-Dickstein, Eric Weiss, Niru Maheswaranathan, and Surya Ganguli.
\newblock Deep unsupervised learning using nonequilibrium thermodynamics.
\newblock In \emph{International conference on machine learning}, pages 2256--2265. PMLR, 2015.

\bibitem[Song et~al.(2020)Song, Sohl-Dickstein, Kingma, Kumar, Ermon, and Poole]{song2020score}
Yang Song, Jascha Sohl-Dickstein, Diederik~P Kingma, Abhishek Kumar, Stefano Ermon, and Ben Poole.
\newblock Score-based generative modeling through stochastic differential equations.
\newblock \emph{arXiv preprint arXiv:2011.13456}, 2020.

\bibitem[Team(2024)]{kolors}
Kolors Team.
\newblock Kolors: Effective training of diffusion model for photorealistic text-to-image synthesis.
\newblock \emph{arXiv preprint}, 2024.

\bibitem[Wang et~al.(2024{\natexlab{a}})Wang, Spinelli, Wang, Bai, Qin, and Chen]{wang2024instantstyle}
Haofan Wang, Matteo Spinelli, Qixun Wang, Xu Bai, Zekui Qin, and Anthony Chen.
\newblock Instantstyle: Free lunch towards style-preserving in text-to-image generation.
\newblock \emph{arXiv preprint arXiv:2404.02733}, 2024{\natexlab{a}}.

\bibitem[Wang et~al.(2024{\natexlab{b}})Wang, Xing, Huang, Ai, Wang, and Bai]{wang2024instantstyleplus}
Haofan Wang, Peng Xing, Renyuan Huang, Hao Ai, Qixun Wang, and Xu Bai.
\newblock Instantstyle-plus: Style transfer with content-preserving in text-to-image generation.
\newblock \emph{arXiv preprint arXiv:2407.00788}, 2024{\natexlab{b}}.

\bibitem[Wang et~al.(2024{\natexlab{c}})Wang, Bai, Wang, Qin, Chen, Li, Tang, and Hu]{wang2024instantid}
Qixun Wang, Xu Bai, Haofan Wang, Zekui Qin, Anthony Chen, Huaxia Li, Xu Tang, and Yao Hu.
\newblock Instantid: Zero-shot identity-preserving generation in seconds.
\newblock \emph{arXiv preprint arXiv:2401.07519}, 2024{\natexlab{c}}.

\bibitem[Wang et~al.(2024{\natexlab{d}})Wang, Darrell, Rambhatla, Girdhar, and Misra]{wang2024instancediffusion}
Xudong Wang, Trevor Darrell, Sai~Saketh Rambhatla, Rohit Girdhar, and Ishan Misra.
\newblock Instancediffusion: Instance-level control for image generation.
\newblock In \emph{Proceedings of the IEEE/CVF Conference on Computer Vision and Pattern Recognition}, pages 6232--6242, 2024{\natexlab{d}}.

\bibitem[Wang et~al.(2024{\natexlab{e}})Wang, Fu, Huang, He, and Jiang]{wang2024ms}
X Wang, Siming Fu, Qihan Huang, Wanggui He, and Hao Jiang.
\newblock Ms-diffusion: Multi-subject zero-shot image personalization with layout guidance.
\newblock \emph{arXiv preprint arXiv:2406.07209}, 2024{\natexlab{e}}.

\bibitem[Xie et~al.(2024)Xie, Chen, Chen, Cai, Lin, Zhang, Li, Lu, and Han]{xie2024sana}
Enze Xie, Junsong Chen, Junyu Chen, Han Cai, Yujun Lin, Zhekai Zhang, Muyang Li, Yao Lu, and Song Han.
\newblock Sana: Efficient high-resolution image synthesis with linear diffusion transformers.
\newblock \emph{arXiv preprint arXiv:2410.10629}, 2024.

\bibitem[Xie et~al.(2023)Xie, Li, Huang, Liu, Zhang, Zheng, and Shou]{xie2023boxdiff}
Jinheng Xie, Yuexiang Li, Yawen Huang, Haozhe Liu, Wentian Zhang, Yefeng Zheng, and Mike~Zheng Shou.
\newblock Boxdiff: Text-to-image synthesis with training-free box-constrained diffusion.
\newblock In \emph{Proceedings of the IEEE/CVF International Conference on Computer Vision}, pages 7452--7461, 2023.

\bibitem[Yang et~al.(2024)Yang, Yu, Meng, Xu, Ermon, and Bin]{yang2024mastering}
Ling Yang, Zhaochen Yu, Chenlin Meng, Minkai Xu, Stefano Ermon, and CUI Bin.
\newblock Mastering text-to-image diffusion: Recaptioning, planning, and generating with multimodal llms.
\newblock In \emph{Forty-first International Conference on Machine Learning}, 2024.

\bibitem[Yang et~al.(2023)Yang, Wang, Gan, Li, Lin, Wu, Duan, Liu, Liu, Zeng, et~al.]{yang2023reco}
Zhengyuan Yang, Jianfeng Wang, Zhe Gan, Linjie Li, Kevin Lin, Chenfei Wu, Nan Duan, Zicheng Liu, Ce Liu, Michael Zeng, et~al.
\newblock Reco: Region-controlled text-to-image generation.
\newblock In \emph{Proceedings of the IEEE/CVF Conference on Computer Vision and Pattern Recognition}, pages 14246--14255, 2023.

\bibitem[Ye et~al.(2023)Ye, Zhang, Liu, Han, and Yang]{ye2023ip}
Hu Ye, Jun Zhang, Sibo Liu, Xiao Han, and Wei Yang.
\newblock Ip-adapter: Text compatible image prompt adapter for text-to-image diffusion models.
\newblock \emph{arXiv preprint arXiv:2308.06721}, 2023.

\bibitem[Yu et~al.(2022)Yu, Xu, Koh, Luong, Baid, Wang, Vasudevan, Ku, Yang, Ayan, et~al.]{yu2022scaling}
Jiahui Yu, Yuanzhong Xu, Jing~Yu Koh, Thang Luong, Gunjan Baid, Zirui Wang, Vijay Vasudevan, Alexander Ku, Yinfei Yang, Burcu~Karagol Ayan, et~al.
\newblock Scaling autoregressive models for content-rich text-to-image generation.
\newblock \emph{arXiv preprint arXiv:2206.10789}, 2\penalty0 (3):\penalty0 5, 2022.

\bibitem[Zhang et~al.(2023)Zhang, Rao, and Agrawala]{zhang2023adding}
Lvmin Zhang, Anyi Rao, and Maneesh Agrawala.
\newblock Adding conditional control to text-to-image diffusion models.
\newblock In \emph{Proceedings of the IEEE/CVF International Conference on Computer Vision}, pages 3836--3847, 2023.

\bibitem[Zhou et~al.(2024)Zhou, Li, Ma, Zhang, and Yang]{zhou2024migc}
Dewei Zhou, You Li, Fan Ma, Xiaoting Zhang, and Yi Yang.
\newblock Migc: Multi-instance generation controller for text-to-image synthesis.
\newblock In \emph{Proceedings of the IEEE/CVF Conference on Computer Vision and Pattern Recognition}, pages 6818--6828, 2024.

\end{thebibliography}
}


\end{document}